\documentclass[11pt,a4paper]{article}
\usepackage{times,latexsym}
\usepackage{url}
\usepackage{graphicx}
\usepackage{booktabs}
\usepackage{comment}
\usepackage{amssymb}
\usepackage{enumitem}
\usepackage{array,multirow}
\usepackage{soul}
\usepackage{color, colortbl}
\usepackage[normalem]{ulem}
\usepackage{makecell}

\usepackage{arabtex}
\usepackage{tipa}
\usepackage[T1]{fontenc}
\usepackage[utf8]{inputenc}

\usepackage[acceptedWithA]{tacl2018v2}

\usepackage{xspace,mfirstuc,tabulary}

\newif\iftaclinstructions
\taclinstructionsfalse 
\iftaclinstructions
\renewcommand{\confidential}{}
\renewcommand{\anonsubtext}{(No author info supplied here, for consistency with
TACL-submission anonymization requirements)}
\newcommand{\instr}
\fi

\iftaclpubformat 

\else

\fi

\title{SentiPers: A Sentiment Analysis Corpus for Persian}

\author{
\fontsize{11pt}{11pt}\selectfont
 \makecell{Pedram Hosseini$^{1,\ast}$, Ali Ahmadian Ramaki$^{2}$, Hassan Maleki$^{3}$,\\Mansoureh Anvari$^{4}$, Seyed Abolghasem Mirroshandel$^4$} 
\\
\fontsize{10pt}{10pt}\selectfont
\makecell{$^{1}$The George Washington University, $^{2}$Ferdowsi University of Mashhad\\ $^{3}$Shahid Beheshti University, $^{4}$University of Guilan\\
{\tt phosseini@gwu.edu, mirroshandel@guilan.ac.ir}
}
}

\newcommand\blfootnote[1]{%
  \begingroup
  \renewcommand\thefootnote{}\footnote{#1}%
  \addtocounter{footnote}{-1}%
  \endgroup
}

\date{}

\begin{document}
\maketitle
\begin{abstract}
\blfootnote{$\ast$ Corresponding Author}
Sentiment Analysis (SA) is a major field of study in natural language processing. With a growing interest in SA over the recent years, there is an increasing need for developing appropriate resources and datasets. In this paper, we outline the entire process of developing an annotated sentiment corpus, SentiPers,\footnote{\scriptsize\url{https://github.com/phosseini/sentipers}\label{github-link}} which covers formal and informal written contemporary Persian. To the best of our knowledge, SentiPers is a unique sentiment corpus with such a rich annotation for Persian. The corpus contains more than 26,000 sentences and benefits from special characteristics such as quantifying the positiveness or negativity of an opinion through assigning a number within a specific range to any given sentence. Furthermore, we present statistics on various components of our corpus as well as the inter-annotator agreement.
\end{abstract}

\section{Introduction}
\label{sec:introduction}
With the rapid growth of media outlets such as forums, blogs, and social networks on the World Wide Web, there are plenty of online resources containing useful opinions and reviews by the customers on various products and services. Feedback and materials generated by customers have been increasingly tapped by individuals and organizations for developing business strategies and improving products and services~\cite{liu2012sentiment}. In other words, “What other people think” has always been of great importance in decision making process~\cite{panglee2008}. In addition, such a rich pool of data can serve as a great resource for academic research. Sentiment analysis (SA) is a major task within the greater field of Natural Language Processing (NLP) and has recently become an increasingly active research area~\cite{liu2012sentiment}. This is a process where opinions on different features of a product, for instance a cell phone or a digital camera, are analyzed to provide an overview of positive or negative sentiments about that product~\cite{liu2012sentiment}. In the field of SA, having access to the appropriate source of data is a necessity for conducting research works such as running various machine learning algorithms. One kind of these resources or datasets used for SA is known as sentiment or opinion corpus.

Most of the current research is focused on developing sentiment corpora for English. Therefore, there is much room and need for research and developing opinion corpora for non-English languages. In particular, there is no sentiment corpus that has been developed for Persian with rich annotation in all levels of analysis. Even among the constructed corpora for English, only a few are publicly available for research and academic work. In addition, in most of these developed sentiment corpora, the polarity detection and analysis is at the document-level or sentence-level and only a few include the third analysis level that is the aspect-level~\cite{liu2012sentiment}.

In this article, we delineate the process of developing a new corpus for Persian called SentiPers. Our corpus is composed of more than 26,000 manually annotated sentences in Persian. One of the important features of SentiPers is the inclusion of both formal (written) and informal (verbal) sentences. In addition, SentiPers rates the polarity of sentences within a range including five numbers to determine the intensity of sentiment orientation. Using such a rating system to determine the polarity may be further used in finding a relation between the sentiment orientation and the number of opinion words in them for any future work. In essence, our corpus consists of the annotation at all three levels including document-level, sentence-level, and entity/aspect level~\cite{liu2012sentiment}.

In the following sections, we initially review the work related to the development of sentiment corpora in different languages in section~\ref{sec:related:work}. In section~\ref{sec:corpus:data}, we explain the process of gathering data for constructing our corpus in detail. Thereafter, we introduce some of the terminologies and concepts of SentiPers that are used in the annotation process in section~\ref{sec:annotation}. The statistics on the corpus and calculation of the inter-annotator agreement are included in section~\ref{sec:statistics}. In section~\ref{sec:challenges}, we discuss some of the challenges we faced in the annotation process. In the end, we highlight the conclusions of this research in section~\ref{sec:conclusion}.

\section{Related Work}
\label{sec:related:work}
In the field of sentiment analysis, possessing a rich and reliable resource is of great importance. There are several sentiment corpora that are publicly available for researchers in this field of study, however, most of these corpora are developed for English meaning sentiment resources and datasets for other languages are rather limited. In this section, we review the most popular opinion mining corpora starting with the ones developed for English followed by the opinion corpora of certain other languages worldwide. We also review some of the sentiment corpora developed for Persian in a separate paragraph.

There has been a number of works in developing sentiment-related resources prior to the year 2000~\cite{wiebe2005annotating}, the time after which the sentiment analysis started to increasingly become one of the most active research areas within NLP~\cite{liu2012sentiment}. In most of these corpora, the sentiment annotation has been done at sentence-level by assigning a sentiment polarity to a sentence~\cite{bethard2004automatic,kim2004determining,yu2003towards} while in some the target words have been additionally annotated in each sentence~\cite{hu2004mining}. The corpus developed by Hu and Liu has the additional feature where each sentence is annotated and the contextual sentiment value is given. The sentences used in this work have been extracted from the online reviews of five consumer electronic devices that include 113 documents spanning 4,555 sentences and 81,855 tokens. MPQA~\cite{wiebe2005annotating} is another sentiment-related corpus that has been extensively used by researchers within the opinion mining community and contains 10,657 sentences in 535 documents. MPQA is mostly composed of news articles and documents manually annotated for opinions and private statements such as beliefs, emotions, sentiments, and speculations. The more recent version of this corpus includes two new annotation types, namely attitude and target annotations. Both of the aforementioned corpora annotate the target words and include the entity and aspect level analysis. Cornell movie review dataset~\cite{pang2002thumbs} is another popular resource for sentiment analysis that includes datasets such as sentiment polarity (document- and sentence-level), sentiment scale, as well as subjectivity. JDPA sentiment corpus~\cite{kessler2010icwsm} is an online resource that contains a wealth of user-generated materials such as blog posts on automobiles~\cite{kessler2010icwsm}. In addition to the various annotation types, JDPA provides examples and statistics on the occurrence and inter-annotator agreement that helps to quantify sentiment phenomena and allows for the construction of advanced sentiment systems. In another research, Twitter has been used for building an opinion mining corpus of 300,000 text posts containing positive, negative, and objective emotions~\cite{pak2010twitter}. In this work, the authors perform statistical linguistic analysis of the corpus and use the collected corpora to build an opinion classification system for microblogging. In addition, they conduct experimental evaluations on a set of real microblogging posts to illustrate that their technique is efficient. The last notable dataset is a resource developed based on the product reviews on Amazon where polarity has been determined using a 1-to-5 scoring system and defining a threshold value for positivity or negativity of the overall rating~\cite{blitzer2007biographies}.

There has been a number of attempts in developing sentiment corpus for non-English languages such as Opinion Corpus in Arabic (OCA) that is composed of 500 movie reviews from Arabic blogs and websites~\cite{rushdi2011oca}. In this work, the reviews have been classified to positive and negative classes and results have been validated through a comparison to the performance of Support Vector Machine and Naïve Bayes algorithms. There is another sentiment corpus developed for Arabic named AWATIF~\cite{abdul2012awatif}. This corpus is a multi-genre corpus of Modern Standard Arabic that is labeled for subjectivity and sentiment analysis at the sentence-level. Another notable SA corpus for non-English languages is ChnSenti-Corp for Chinese~\cite{tan2008empirical} that consists of 1,021 documents in three domains namely education, movie, and housing where each of these categories has positive and negative documents. MLSA is a publicly available multi-layered (document, sentence, phrase, and expression levels) annotated sentiment corpus for German-language~\cite{clematide2012mlsa}. The construction of this corpus is based on the manual annotation of 270 German-language sentences. Average pairwise agreement and Fleiss’ multi-rater Kappa~\cite{fleiss1981statistical} are used to calculate the reliability of this sentiment corpus. There have been some multilingual corpora developed for SA as well, most notably NTCIR that includes Japanese, English, traditional Chinese, and simplified Chinese where the process of annotation and evaluation approaches has been discussed for each language~\cite{seki2008overview}. USAGE is another fine-grained multilingual sentiment corpus that includes both German and English. This resource contains the annotation of the product reviews selected from Amazon with both aspects and subjective phrases~\cite{klinger2014usage}.

Unlike the significant number of rich corpora developed for English, there has not been much sentiment corpora developed for Persian. In addition, most of these Persian corpora have been labeled only at document-level or sentence-level. In this paragraph, we introduce the available sentiment corpora for Persian. For evaluating a LDA-based algorithm for sentiment classification, a collection of user reviews were extracted from three domains including cell phones, digital cameras, and hotels from some Persian e-shopping websites~\cite{shams2012non}. The polarity of the reviews in this collection has been assigned manually. Finally, for each domain, 200 positive and 200 negative reviews were chosen for evaluating the proposed method. In another research, for testing a Persian sentiment analyzing method, a dataset has been generated composed of 511 positive and 509 negative online customer reviews in Persian from some brands of cell phone products. Two annotators labeled these reviews manually~\cite{bagheri2014persian,saraee2013feature}. Another dataset has been created named BS Data containing user reviews from a Persian website, mobile.ir. This dataset is composed of a total number of 263 positive and negative reviews~\cite{basiri2014framework}. In another study, a dataset is collected from a Persian website named hellokish on the hotel domain. This dataset contains 1,805 negative and 4,630 positive reviews. Each review has some attributes including an opinion about the hotel, its date, and writer as well~\cite{alimardani2015opinion}. There are also some other collections of Persian reviews that have been used in studies on sentiment analysis in Persian~\cite{golpar2015feature,hajmohammadi2013svm}.

\section{Corpus Data}
\label{sec:corpus:data}
The first and one of the most important steps in developing a corpus is selecting the appropriate data source. The data used in the construction of SentiPers is extracted from a website named Digikala.\footnote{\scriptsize\url{https://www.digikala.com/}} Digikala is the most widely-used website in online shopping of electronic products (e.g., cell phones, printers, digital cameras) in Iran, holding a similar place as Amazon in the United States. In addition to online shopping, thousands of individuals visit this website every day in order to review various aspects of a range of products. These reviews are a useful resource for visitors in making the optimum choice that meets their needs. All these characteristics make Digikala one of the most popular online shopping websites in Iran.

Among all the resources we could possibly use for developing our corpus, Digikala stood out as the most suitable candidate due to the following reasons. First, there are some Persian websites with a noticeable number of opinions stated by various individuals, but for the specific domain that we chose to work on, electronic products, Digikala offers some unique characteristics. For instance, the number of visitors of the website and more importantly the number of people who review different products are substantial. Furthermore, Digikala has been chosen as the best electronic shop in Iran several times and is trusted widely by a large portion of the Iranian population. Aside from these reasons, an additional factor that makes Digikala an appropriate choice is the fact that opinionated written materials of this website can be organized into two distinct sections that are formal and informal natural languages. For each product available on the website, there is a section named \emph{criticizing and discussion} that covers the technical opinion of an expert about a product. The language of this section is formal. The following sentences are examples of formal Persian texts:

{\footnotesize\setfarsi\novocalize \<.trA.hy w sAxt ayn gw^sy bsyAr `Aly w kyfyt t.sAwyr\\ An by\hspace{0ex}n.zyr ast>} /tærâhi væ sâxt-e in gôshi besiâr âli væ keifiæt-e tæsâvir niz dær ân bi næzir æst/ (The design and manufacturing of this phone is great and pictures are of excellent quality).

{\footnotesize\setfarsi\novocalize \<ayn dwrbyn yky az bhtryn m.h.swlAt ^srkt swny bh\\ .hsAb my\hspace{0ex}Ayd w dArAy wy^zgy\hspace{0ex}hAy mn.h.sr bh frdy ast>} /in dôrbin yeki æz bærtærin mæhsôlât-e sherkæt-e Sony be hesâb miâyæd væ dârâye vijegi hâye monhæser be færdi æst/ (This camera considers one of the best products of Sony Corporation presenting some unique features).

The \emph{general reviews} and \emph{critical reviews} sections of the website are managed by users who are not necessarily experts. The language used in these two sections is typically informal as opposed to the section written by the experts. In such an informal text, the order of elements of a sentence may be slightly modified. For example, a simple pattern in formal Persian is: Subject + Object + Verb and an example for this pattern is: {\footnotesize\setfarsi\novocalize \<mn `aly rA dydm>} /mæn æli râ didæm/ (I saw Ali). Informal sentences, however, do not necessarily follow this formal pattern. For instance, in the sentence: {\footnotesize\setfarsi\novocalize \<dydm^s `aly rw>} /didæmesh æli ro/ (I saw Ali), the pattern is instead: Verb + Subject + Object. Additionally, the structure of the words may be subject to change in informal sentences. For instance, {\footnotesize\setfarsi\novocalize \<`alyw dydm^s>} /ælio didæmesh/ (I saw Ali.), in fact, the word {\footnotesize\setfarsi\novocalize \<`alyw>} /ælio/ is the combination of two words {\footnotesize\setfarsi\novocalize \<`aly>} /æli/ and {\footnotesize\setfarsi\novocalize \<rA>} /râ/ where due to the nature of informal language these two words are combined to a single word. Following sentences are further examples of informal Persian:

{\footnotesize\setfarsi\novocalize \<ayn dwrbyn bh n.zrm `Alyh w `aksAy^y kh mygyrh wAq`A\\ bAkyfyth>} /in gôshi be næzæræm âlieh væ æxâyi ke migireh vâqeæn bâ keifiæteh/ (In my opinion this phone is great and the pictures that the phone takes really have good quality).

{\footnotesize\setfarsi\novocalize \<ayn dwrbyn wAq`A m.h^srh. tw xryd^s yk dr.sd hm ^sk\\ nknyn.>} /in dôrbin vâqeæn mæhshæreh. Tô xæridesh yek dærsæd hæm shæk nækonin/ (This camera is really amazing. Do not hesitate to buy it).

\begin{figure}[]
\centering
\includegraphics[scale=0.53]{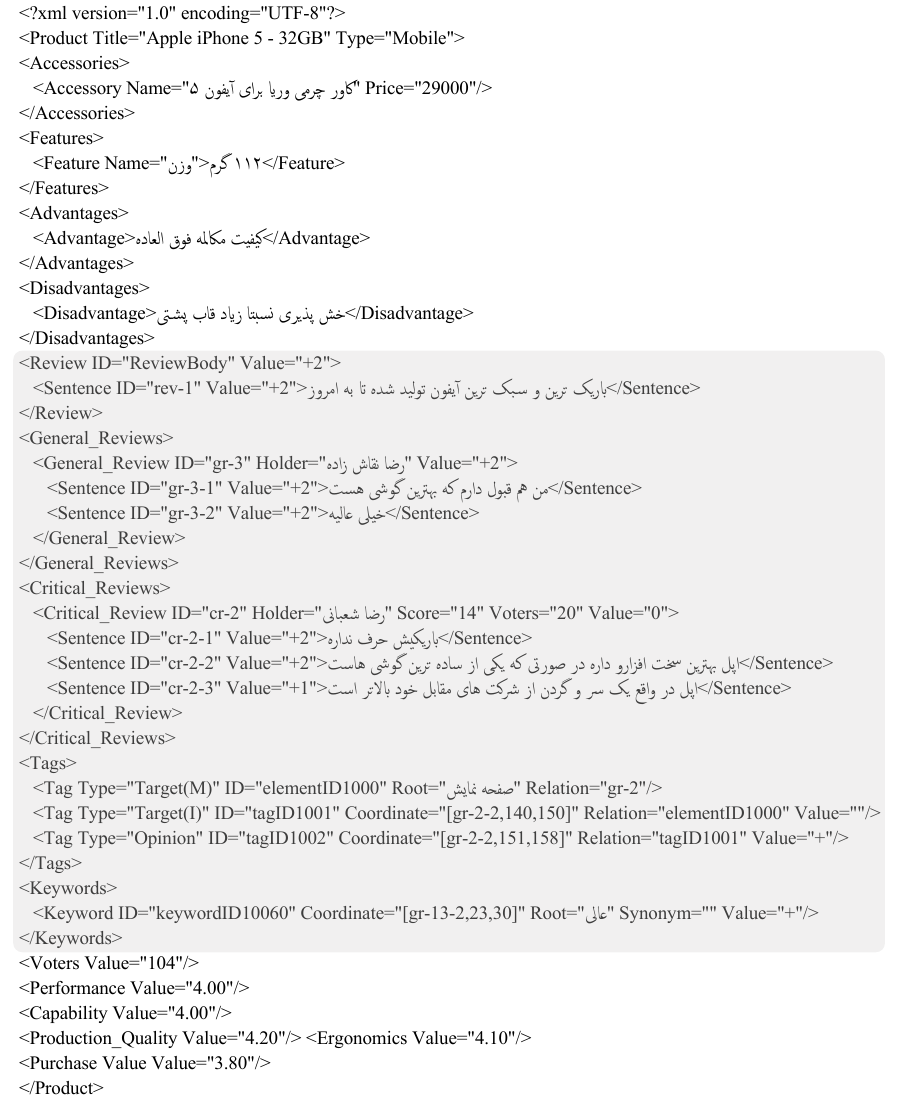}
\caption{\label{fig:data}Structure of a sample XML file in SentiPers}
\end{figure}

Once Digikala was identified as the best available data source, the website was thoroughly crawled.\footnote{Based on the terms and conditions of Digikala, the information of the website is allowed to be used for non-commercial activities with referring to Digikala.} In the next step, the HTML pages of products gained from crawling Digikala were converted to XML files. Figure~\ref{fig:data}, shows the structure of a sample XML file.

Generally, each XML file consists of complete information about a specific product. One of the main parts of the XML file consists of three elements named \emph{Review}, \emph{Critical\_Reviews}, and \emph{General\_Reviews}. The reason that we separated opinions into three different categories is that these opinions may have important differences in comparison to one other. They may be either formal or informal or they may be written by either an expert or a non-expert user. The text body of all the three parts in XML files is divided into sentences. Each sentence has a unique ID. This ID specifies the order of the sentence among all the sentences of an opinion. In addition, it shows which one of the three parts the sentence belongs to. A collection of one or more sentences then form an opinion or a document.

\section{Annotation process}
\label{sec:annotation}
The next step after the preparation of the raw XML files is annotating the corpus. Prior to going through the annotation process, certain concepts related to the SentiPers must be explained in more detail. There are four annotators contributing to our corpus. These annotators were trained by reviewing an annotation guideline as well as the annotation of several sample documents of the corpus. In addition, all of the annotators are Persian native speakers with proper knowledge and understanding of Persian grammar as well as some background knowledge of sentiment analysis. In the end, an experienced annotator reviewed all of the annotated documents.

\subsection{Types of tags}
\subsubsection{Target and opinion words}
There are two types of tags namely \emph{Target} and \emph{Opinion} words in our corpus. Target word is an entity or an aspect of an entity described by one or more opinion words~\cite{liu2012sentiment}. In the rest of the paper, we use the short forms of target and opinion for target word, and opinion word, respectively. We review an example here in order to clarify the meaning of these two types of tags. Consider the sentence {\footnotesize\setfarsi\novocalize \<ayn gw^sy `Aly ast>} /in gôshi âli æst/ (This phone is great). In this sentence, the word phone is a target word that has been described by great as an opinion word.

\subsubsection{Keyword}
\emph{Keywords} are another type of tag that are somehow similar to opinion word since they may have a $+$ or $-$ sense. Keyword has two specific usages. First, there are some cases that even though a sentence does have a sense, but the annotator can not use a pair of opinion and target words in order to select the words that contribute to the sense of the sentence. For instance, consider the sentence: {\footnotesize\setfarsi\novocalize \<ayn gw^sy xwb nyst>} /in gôshi xôb nist/ (This phone is not good). It is clear that the sentence has a negative sense about the phone entity. Based on our guideline, if the annotator wants to annotate the sentence using opinion and target, the only possible way is choosing \emph{phone} as the target and \emph{is not good} as opinion. However, to make our corpus useful for applying various algorithms similar to what has been done in the composition model in~\cite{moilanen2007sentiment}, we came to the conclusion that separate selection of sensed words as keywords works better in comparison to selecting a pair of opinion and target. In the example mentioned earlier, we annotate two keywords: \emph{good} with a positive sense and \emph{is not} with a negative sense. As a result, in further research, for instance, by analyzing the composition of these two keywords we can easily conclude that the sense of the sentence is negative because the negative verb comes right after a positive adjective.

Another usage for the keyword is in annotating the strength of the polarity. In some cases, certain words in the sentence directly illustrate the strength of the polarity in the sentence. For example, in {\footnotesize\setfarsi\novocalize \<ayn gw^sy `Aly ast>} /in gôshi âli æst/ (this phone is great), the word great clearly shows that the degree of positiveness is strong. However, there are some cases that words with polarity may not contribute to the strength of the polarity alone. For instance, in the sentence {\footnotesize\setfarsi\novocalize \<ayn gw^sy a.slA xwb nyst>} /in gôshi æslæn xôb nist/ (This phone is not good at all.), the reviewer does not only believe that the phone is not good, but he emphasizes on his comment by using at all. In such cases, words like at all can be annotated as keywords and they are a kind of intensifier. In similar conditions and in further processing, keywords may help us figure out why a specific polarity is assigned to a sentence. 

Some opinions in sentences are annotated as keywords as well. The reason is that in future work, these opinionated keywords may be useful in building a sentiment lexicon. As a result, part of the keywords may be a subset of opinions.

\subsection{Polarity assignment}
In addition to selecting the appropriate target and opinion words in a sentence, assigning a sentiment polarity to each document and sentence is important as well. The polarity that has been assigned to each document and sentence is a number from the set $\{-2, -1, 0, +1, +2\}$ that shows the sentiment orientation of the sentence by $-2$ being the most negative and $+2$ being the most positive. The value $0$ shows that polarity of the sentence is neutral. In the following, there are several examples of sentences with different polarities:

{\footnotesize\setfarsi\novocalize \<gw^sy\hspace{0ex}ay kh hfth py^s xrydm fAj`ah hst w kAmlA nA\\ amydm krd>} /gôshi ke hæfte-ye pish xæridæm fâjee hæst væ kâmelæn nâ omidæm kærd/ (The cell phone that I bought last week is a disaster and made me totally disappointed). [polarity: -2]

{\footnotesize\setfarsi\novocalize \<kyfyt .sf.hh nmAy^s tlwyzywn xyly bdh w nmytwnm\\ t.hml^s knm>} /keifiæt-e sæfhe næmâyesh-e televizion xeili bædeh væ æslæn nemitônæm tæhæmmolesh konæm/ (The quality of TV screen is very bad and I cannot stand it at all). [polarity: -2]

{\footnotesize\setfarsi\novocalize \<bAtry gw^sy xwb kAr myknh, hr^cnd sAyz .sf.hh nmAy^s\\ mnAsb nyst w mn ayn gw^sy rw dwst ndArm>} /bâtri-ye gôshi xôb kâr mikoneh, hærchænd sâyz-e sæfhe næmâyesh monâseb nist væ mæn in gôshi ro dôst nædâræm/ (The cell battery works properly; however, size of the screen is not appropriate and I do not like the phone). [polarity: -1]

{\footnotesize\setfarsi\novocalize \<agr^ch kyfyt t.sAwyry kh bA ayn dwrbyn grfth ^sdh\\ tqrybA xwb ast amA rzwlw^sn An rA.dy\hspace{0ex}knindh nyst >} /keifiæt-e tæsâviri ke bâ in dôrbin gerefte shodeh tæghribæn xôb æst, æmmâ resolution ân râzi konændeh nist/ (Even though the quality of the pictures taken by this camera is relatively good, the resolution is not that satisfying). [polarity: -1]

{\footnotesize\setfarsi\novocalize \<mn ayn pryntr rw mAh py^s xrydm. dr hr dqyqh byst\\ .sf.hh rw ^cAp myknh w rng^s sfydh>} /mæn in printer ro mâhe pish xæridam. dær hær dæqiqeh bist sæfhe ro châp mikoneh væ rængesh sefideh/ (I bought this printer last month. It prints twenty pages a minute and the printer”s color is white). [polarity: 0]

{\footnotesize\setfarsi\novocalize \<kyfyt t.swyr xwbh dr .hAly kh wydywy grfth ^sdh bA\\ ayn dwrbyn ^cndAn xwb nyst>} /keifiæt-e tæsvir xôbeh dar hâlikeh keifiæt-e video-ye gerefteh shode bâ in dôrbin chændân xub nist/ (The picture quality is good while the quality of the video taken by the camera is not that good). [polarity: 0]

{\footnotesize\setfarsi\novocalize \<m.srf anr^zy ayn gw^sy xwbh. dr mjmw`a az^s rA.dym>} /mæsræf-e energy-ye gôshi xôbeh. Dær mæjmô æzæsh râziæm/ (The phone power consumption is good. Generally, I‘m happy with it). [polarity: +1]

{\footnotesize\setfarsi\novocalize \<sAmswng glksy mn hmyn alAn bh dstm rsyd. xwb kAr\\ myknh w dwst^s dArm>} /Samsung Galaxy-ye mæn hæmin ælân be dæstæm resid. xôb kâr mikoneh væ dôstesh dâræm/ (I just received my Samsung Galaxy. It works fine and I like it). [polarity: +1]

{\footnotesize\setfarsi\novocalize \<ayn gw^sy wAq`aA ^sgft\hspace{0ex}angyzh, kyfyt t.sAwyr^s xyly\\ `aAlyh>} /in gôshi vâgheæn shegeft ængizeh. keifiæt-e tæsâviresh xeili âlieh/ (The phone is really amazing. The quality of its pictures is excellent). [polarity: $+2$]

{\footnotesize\setfarsi\novocalize \<rzwlw^sn wAq`aA xwbh. andAzh .sf.hh nmAy^s `aAlyh w hy^c\\ ^cyz bdy dr mwrd^s wjwd ndArh>} /resolution vâgheæn xôbeh. ændâzeye sæfhe næmæyesh âlieh væ hich chiz-e bædi dær moredesh vôjôd nædâreh/ (The resolution is really good. The screen size is great and there is really nothing bad about it). [polarity: +2]

\subsection{Annotation tool and corpus availability}
\begin{figure}[]
\centering
\includegraphics[scale=0.23]{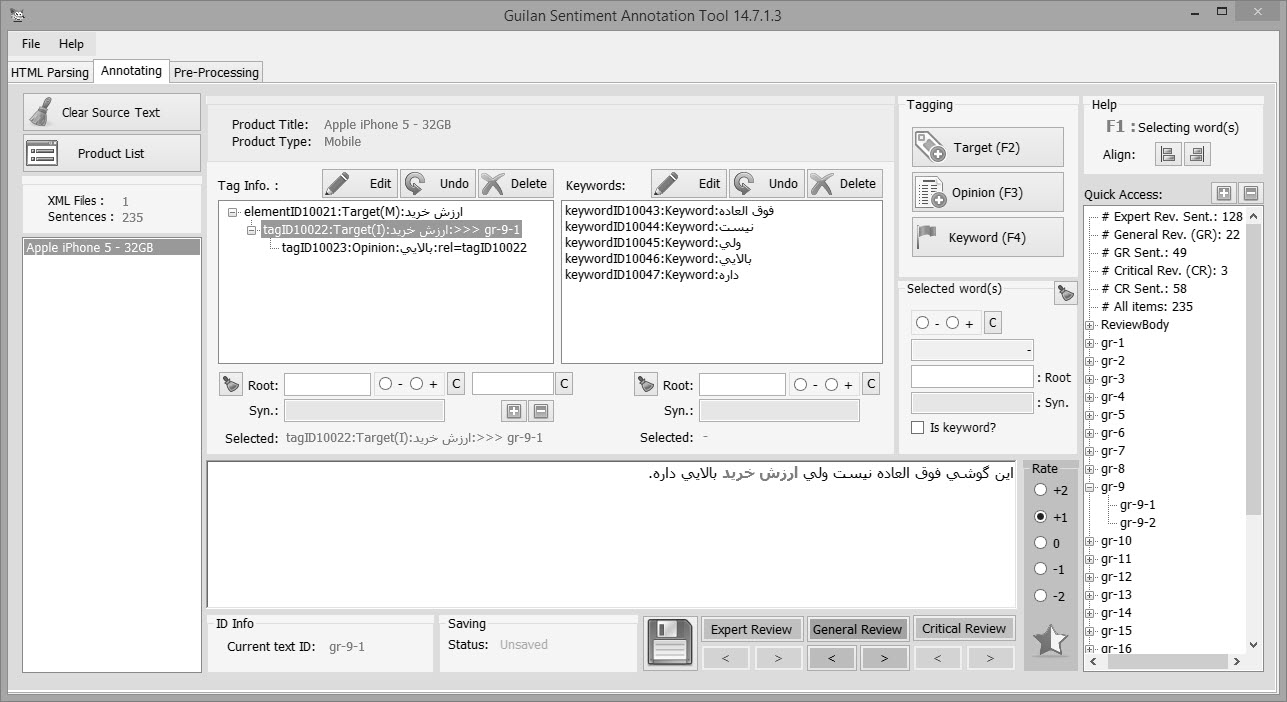}
\caption{\label{fig:software}The snapshot of the software implemented for annotation process}
\end{figure}

For the annotation process, we developed an annotation software. The software is implemented specifically for annotation, measuring the statistics related to SentiPers (e.g., number of words and tokens, number of sentences, and so on), as well as the inter-annotator agreement. Aside from the annotation process, the software includes an editor for receiving the information from HTML pages using XPath~\cite{berglund2003xml}. This editor helps us to find those HTML tags that contain the required information for building the XML files. A snapshot of the software environment is shown in Figure~\ref{fig:software}. It is important to mention that our corpus, SentiPers, is publicly available for research and noncommercial activities.

\section{Corpus statistics}
\label{sec:statistics}
In this section, we present the statistics of SentiPers. The process of calculating Inter-Annotator Agreement (IAA) will be discussed in the following section. Table~\ref{tab:word:statistics}, shows the most important statistics of our corpus.

\begin{table}[]
\centering
\small
\begin{tabular}{cc}
\toprule
Title & Count \\ \midrule
XML Documents                      & 270            \\ 
Sentences                          & 26,767         \\ 
Tokens                             & 515,387        \\ 
Unique Words                       & 17,635         \\ 
Opinion Words                      & 26,996         \\ 
Target Words                       & 21,375         \\ 
Average Length of Sentences (Word) & 19.25          \\ \bottomrule
\end{tabular}
\caption{\label{tab:word:statistics}General statistics of SentiPers}
\end{table}

The number of opinion words categorized by their polarity is also illustrated in Table~\ref{tab:polarity}. Table~\ref{tab:statistics}, shows the count of each type of product among XML documents that have been annotated. As shown in the table, cell phone, digital camera, camcorder, and tablet are the most frequent types of commodities among all products in our corpus.

\begin{table}[]
\centering
\small
\begin{tabular}{cccc}
\toprule
Polarity $\rightarrow$ & Positive & Neutral  & Negative     \\ \midrule
Opinion Word           & 21,471                & 1,661                 & 3,864                 \\
Sentence               & 12,921                & 11,353                & 2,678                 \\ \bottomrule
\end{tabular}
\caption{\label{tab:polarity}The number of opinion words and sentences for different sentiment polarities}
\end{table}

\begin{table}[]
\centering
\small
\begin{tabular}{cc}
\toprule
\multicolumn{1}{c}{Product} & Count \\ \midrule
Cell Phone                             & 72             \\ 
Digital Camera                         & 65             \\ 
Camcorder                              & 37             \\ 
Tablet                                 & 20             \\ 
Notebook                               & 17             \\ 
Printer                                & 13             \\ 
Computer                               & 12             \\ 
Music Player                           & 10             \\ 
TV                                     & 10             \\ 
Game Console                           & 7              \\ 
Scanner                                & 7              \\ 
\emph{Total}                         & \emph{270}            \\ \bottomrule
\end{tabular}
\caption{\label{tab:statistics}The document count of each type of product}
\end{table}

\subsection{Inter-annotator agreement}
Because of the subjective nature of manual annotation, calculating the agreement among the annotators is important. There are certain measures that can be implemented in the calculation of the IAA. Some of the well-known measures are Fleiss’s K, Cohen’s Kappa, Cronbach’s Alpha, and Krippendorff’s Alpha~\cite{hayes2007answering}. In the two following subsections, we calculate the agreement among the annotators in selecting different types of tags in the sentences and assigning sentiment polarities to sentences.

\subsubsection{Agreement for polarity assignment}
In order to calculate the IAA for the assigned polarity of the sentences, we used Fleiss‘s kappa measure~\cite{fleiss1973equivalence}. Fleiss‘s kappa is a proper measure here because the values of polarities assigned to sentences are of the nominal type. In Fleiss‘s kappa formula, three categories namely including positive, neutral, and negative are considered. The result of the agreement for polarity assignment is shown in Table~\ref{tab:agreement}.

\subsubsection{Agreement for tags}
Regarding the calculation of the agreement between the annotators for annotated target and opinion words, there are some points that should be mentioned. First of all, since there is no guarantee that the set of target and opinion words annotated by the annotators will be the same, we are not able to use known measures such as Fleiss‘s kappa for calculating the agreement here~\cite{wiebe2005annotating}. In other words, the annotated target and opinion words here are not of the nominal type and there is no fixed number of categories in each sentence for these tags while kappa measures are more suitable for nominal or categorical values~\cite{carletta1996assessing,fleiss1971measuring}. Consequently, for measuring the agreement for the identified target and opinion words by the annotators we used the same method that has been used for measuring agreement for text anchors in~\cite{wiebe2005annotating}. Letting A and B be the sets of anchors annotated by annotators \emph{a} and \emph{b}, the idea behind this method is based on measuring what proportion of A was also marked by b using formula~\ref{formula:agr}.

\begin{equation}
\label{formula:agr}
\mbox{\textit{agr}}(a||b)= \frac{|A\mbox{ \textit{matching} }B|}{|A|}
\end{equation}

It is also necessary to mention that the level of reliability of the IAA rate may be different in various types of corpora. As a result, the IAA may be better to be interpreted and judged based on the type of the annotation task and its level of difficulty in the annotation process. The result of the IAA for annotated tags is shown in Table~\ref{tab:agreement}.

\begin{table}[]
\centering
\small
\begin{tabular}{cc}
\toprule
\multicolumn{1}{c}{Task} & Agreement (\%) \\ \midrule
\multicolumn{1}{c}{Polarity Assignment} & 63.15          \\\
\emph{Opinion} Word Annotation & 67.60 \\
\emph{Target} Word Annotation & 62.60 \\\bottomrule         
\end{tabular}
\caption{\label{tab:agreement}Inter-Annotator Agreement study results}
\end{table}

\section{Challenges}
\label{sec:challenges}
In the following paragraphs, we discuss some of the challenges faced during the annotation process.

On certain occasions, opinion holders do not directly state their opinions about the entities and the features of the entities. For example in the sentence: {\footnotesize\setfarsi\novocalize \<bAtry^s hm m_tl .trA.hy^s a.slA xwb nyst>} /bâtrish hæm mesl-e tærâhish æslæn xôb nist/ (The battery is not good at all, just like the design), the opinion holder is talking about the feature battery of the phone saying this feature is not good at all. At the same time, he thinks that the feature design is not good either; however, this second opinion is not stated directly. In such cases, selecting pairs of target and opinion words is difficult even for a human annotator.

Assigning the correct sense of a sentence regardless of the number of positive or negative words could be a challenging task for human annotators in certain cases. For example, in the sentence {\footnotesize\setfarsi\novocalize \<gw^sy xyly xwb nyst. kyfyt `aks xwby hm\\ ndArh. wly dwst^s dArm>} /gôshi-e xeili xôbi nist. keifiæt-e æks-e xôbi hæm nædâræd. væli dôstæsh dâræm/ (The phone is not very good. The quality of its picture is not good. However, I like it.), even though the opinion holder uses opinions such as is not very good for the entity phone, but at the end of the sentence the opinion is stated directly as positive. In such cases, the existence of an opinion antithetical to another opinion in a sentence makes it difficult for the annotator to assign a polarity to the sentence.

Selecting the correct reference target for an opinion word is important and challenging at times. There are sentences with a certain degree of ambiguity where recognizing the reference of an opinion is not simple. For example, in the sentence {\footnotesize\setfarsi\novocalize\<ayn dwrbyn bh tknwlw^zy py^srfth\hspace{0ex}ay mjhz ^sdh kh my\hspace{0ex}twAn AnrA\\ az jdydtryn\hspace{0ex}hA bh .hsAb Awrd>} /in dôrbin be technology pishræftehi mojæhæz shodeh keh mitævân ân ra æz jædidtærin hâ be shomâr âværd/ (This camera is equipped with advanced technology that can be considered as the latest one). It is not clear whether the opinion holder uses the adjective {\footnotesize\setfarsi\novocalize \<jdydtryn>} /jædidtærin/ (the latest) for the word camera or for the technology. This issue is even more challenging in informal Persian where there is not always a structured written text.

\section{Summary and future work}
\label{sec:conclusion}
In this paper, the process of developing a sentiment corpus comprised of formal and informal contemporary Persian was outlined. We reviewed the structure of documents, the process of annotating these documents by annotators, and addressed some of the challenges faced during the annotation process. In the end, the statistics related to SentiPers such as the number of words and sentences as well as the inter-annotator agreement were presented.

Considering the rich characteristics of SentiPers, this corpus is a unique annotated sentiment resource for researchers interested in working on Persian and in the area of sentiment analysis. There are three specific features that make SentiPers unique compared to existing Persian sentiment corpora. First, SentiPers consists of more than 26,000 sentences that is far more than the number of sentences of other Persian sentiment corpora and even sentiment corpora in languages other than Persian. In addition, SentiPers has been annotated in three different levels including document-, sentence-, and aspect levels unlike other Persian opinion corpora that are annotated at either document- or sentence-level. Moreover, our corpus is publicly available for research and non-commercial activities as well.

As for future work, we are going to expand SentiPers to other domains such as news, politics, and sport as well. Additionally, we aim to operate various machine learning algorithms including deep learning on SentiPers, and evaluate the accuracy of these algorithms.

\bibliography{ref}
\bibliographystyle{acl_natbib}




\end{document}